\documentclass[runningheads]{llncs}
\usepackage[T1]{fontenc}
\usepackage{graphicx,verbatim}
\usepackage{graphicx}
\usepackage{subcaption} 
\usepackage{graphicx}
\usepackage{booktabs}
\usepackage{multirow}
\usepackage{array}
\usepackage{multirow}
\usepackage[table,xcdraw]{xcolor}
\usepackage{amsmath,amsfonts,amssymb}
\usepackage{amsfonts}

\begin{document}
\title{MOSCARD - Causal Reasoning and De-confounding for Multimodal Opportunistic Screening of Cardiovascular Adverse Events}
\titlerunning{MOSCARD}

\author{Jialu Pi\inst{1} \and
Juan Maria Farina\inst{2}\and
Rimita Lahiri\inst{3} \and
Jiwoong Jeong\inst{1} \and
Archana Gurudu\inst{3} \and
yung-Bok Park\inst{5} \and
Chieh-Ju Chao\inst{4} \and
Chadi Ayoub\inst{2} \and
Reza Arsanjani\inst{2} \and
Imon Banerjee\inst{1,3}} 

\authorrunning{J. Pi et al.}


%
\institute{
Department of Data Science \& Engineering, Arizona State University, Tempe, AZ, USA
\and
Department of Cardiovascular Medicine, Mayo Clinic, Scottsdale, AZ, USA
\and
Department of Radiology, Mayo Clinic, Phoenix, AZ, USA
\and
Department of Cardiovascular Medicine, Mayo Clinic, Rochester, MN, USA
\and 
Catholic kwandong university international st. mary’s hospital, South Korea
\\
\email{\{jialupi, jjeong35\}@asu.edu}, 
\email{\{argurudu14, hyungbok7\}@gmail.com}, 
\email{\{Farina.JuanMaria, Lahiri.Rimita, Chao.ChiehJu, Ayoub.Chadi, Arsanjani.Reza, Banerjee.Imon\}@mayo.edu}
}

\maketitle              
\begin{abstract}
Major Adverse Cardiovascular Events (MACE) remain the leading cause of mortality globally, as reported in the Global Disease Burden Study 2021. Opportunistic screening leverages data collected from routine health check-ups and multimodal data can play a key role to identify at-risk individuals. Chest X-rays (CXR) provide insights into chronic conditions contributing to major adverse cardiovascular events (MACE), while 12-lead electrocardiogram (ECG) directly assesses cardiac electrical activity and structural abnormalities. Integrating CXR and ECG could offer a more comprehensive risk assessment than conventional models, which rely on clinical scores, computed tomography (CT) measurements, or biomarkers, which may be limited by sampling bias and single modality constraints. We propose a novel predictive modeling framework - MOSCARD, multimodal causal reasoning  with co-attention to align two distinct modalities and simultaneously mitigate bias and confounders in opportunistic risk estimation. Primary technical contributions are -  (i) multimodal alignment of CXR with ECG guidance; (ii) integration of causal reasoning; (iii) dual back-propagation graph for de-confounding. Evaluated on internal, shift data from emergency department (ED) and external MIMIC datasets, our model outperformed single modality and state-of-the-art foundational models - AUC: 0.75, 0.83, 0.71 respectively. Proposed cost-effective opportunistic screening enables early intervention, improving patient outcomes and reducing disparities\footnote{Code will be available at \url{https://github.com/OrchidPi/MOSCARD}}.


\end{abstract}
%
%
%
\section{Introduction}
Major adverse cardiovascular events (MACE) are among the leading causes of death worldwide \cite{gbd2018global}. Opportunistic screening refers to identifying individuals at risk of MACE during routine health check-ups leveraging multimodal data to provide timely intervention for high and medium risk individual. Although modern clinical practice is highly dependent on the synthesis of information from multiple heterogeneous data sources, traditional predictive models rely on clinical data~\cite{PREISS2015613,LLOYDJONES200420} or CT-based imaging biomarkers~\cite{budoff2018ten,williams2019coronary,suh2015prognostic}. Combination of heterogeneous high-dimensional data is still a substantial technical challenge in healthcare AI due to data complexity, noise, semantic interoperability, bias, and ethical dilemmas~\cite{acosta2022multimodal}. Recent advancements with the vision-language models (e.g., MedClip) remains limited in medical domain due to constraints of the modalities (primarily text and image) and the inherent biases present~\cite{lee2023survey}. Furthermore, these models often exhibit reduced performance in diverse populations due to population bias based on selective sampling~\cite{damen2019performance,Brindle838}. This bias occurs when training data lack sufficient diversity, resulting in shortcut learning and consistent inaccuracies between underrepresented groups ~\cite{BANERJEE2023842,behar2023generalizationmedicalaiperspective,brown2023detecting}. 

While Chest X-ray (CXR) is particularly useful for detecting signs of chronic conditions that contribute to cardiac disease, 12-lead electrocardiogram (ECG) provides a direct assessment of the electrical activity of the heart, detecting arrhythmia, ischemic changes, and structural abnormalities. CXR integrated with 12-lead ECG captured during the same visit could represent a more holistic view of a patient's cardiac health. For opportunistic screening, this combination could be especially useful in settings where resources or access to more sophisticated imaging like echocardiography, CT, magnetic resonance imaging (MRI) are limited. Existing AI models, while effective, often learn demographic biases, leading to disparities in disease prediction. A recent review on `fair' AI model~\cite{correa2022systematic} highlights three key methodologies - fair dataset curation, unbiased representation learning, and adversarial debiasing. Achieving fair dataset by ensuring equal representation of all sensitive attributes can be challenging when data is collected from the real-world distribution. Adversarial debiasing has been shown to reduce racial bias without the accessing balanced dataset~\cite{CORREA2024104548}, however, its complexity limits its effectiveness, as it typically targets only single confounder, failing to model interdependent factors such as age and comorbidities. We propose a multimodal causal reasoning framework - \textbf{MOSCARD} to address the impact of bias and confounders for opportunistic screening for composite short (6 months) and long (5 years) term MACE. The proposed model leverages the posterior-anterior view (PA) of CXR images, and 12-lead ECGs to learn `fair' representation while modeling causality to calculate the probabilistic risk score. 

Our primary contributions are - (i) novel co-alignment framework to integrate the knowledge from ECG data to guide the embedding space alignment for CXR, (ii) remove the confounder effect (selection bias) for CXR and ECG data featurization through confusion loss, (iii) apply causal reasoning to model the causal relationship between comorbidities and the targeted MACE risk (e.g. Obesity → Insulin Resistance → Dyslipidemia → Atherosclerosis → CVD Risk). MOSCARD is compared against the single modality and state-of-the-art multimodal deep learning models~\cite{wang2022medclip,li2021align} without debiasing and causal intervention across multiple datasets (including MIMICIV~\cite{johnson2020mimic}) with and without the domain shift.

\section{Methods}
\begin{figure}[tb!]
    \centering
    \includegraphics[width=0.9\linewidth]{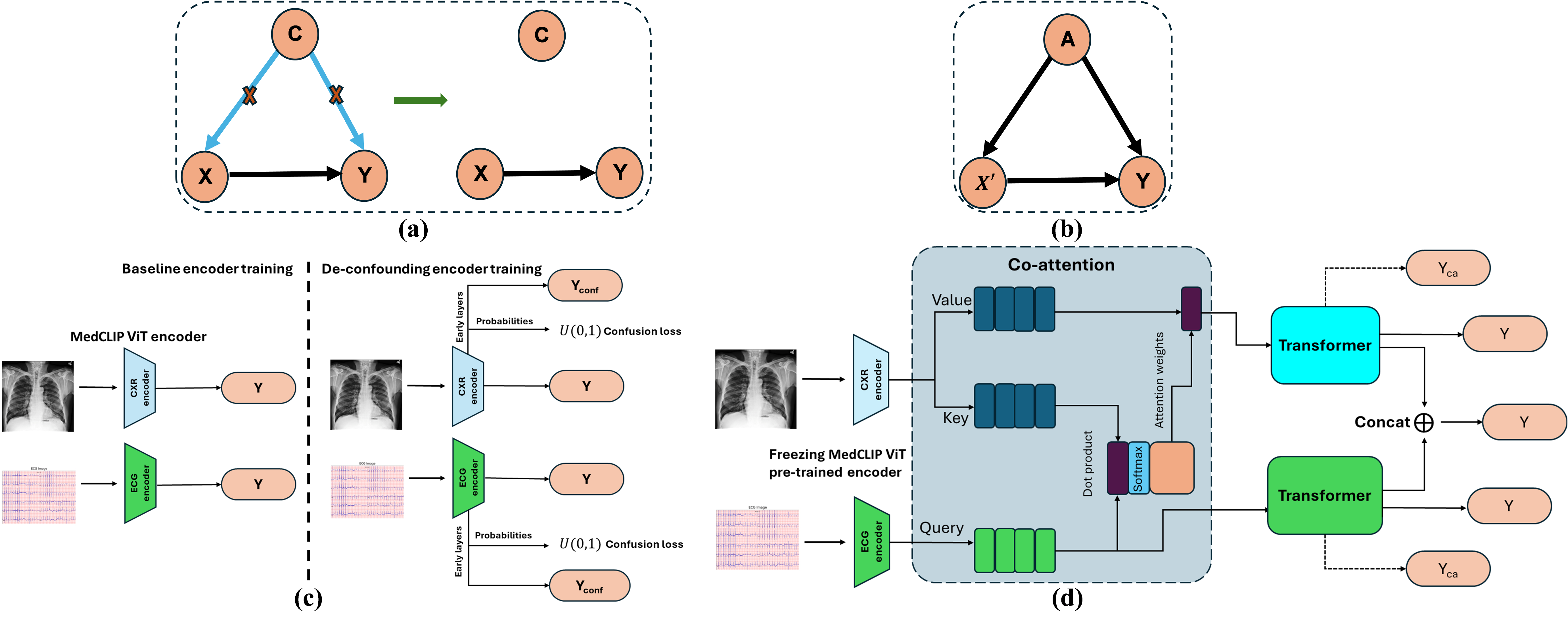}
 \caption{\footnotesize Proposed MOSCARD architecture and de-confounding causal reasoning graph, input $X$, task label $Y$, causal factor $A$, confounder $C$, directed edges for causal confounder relations: (a) Step 1 - single modality encoder training with confusion loss; (b) Step 2, multimodal learning with co-attention and SCM; (c) Step 1 training – Single modality; (d) Step 2 Multimodal training with co-attention and causal intervention.}
\label{fig:multimdal framework}
\end{figure}

We proposed a two step multimodal framework - MOSCARD, where in step 1 we train the single modality encoders with confusion loss to remove  confounder effect (Fig.~\ref{fig:multimdal framework} a) and in step 2 we apply multimodal learning with co-attention (Fig.~\ref{fig:multimdal framework} b). For modeling causal relationships in step 2, we applied a Structural Causal Model (SCM) and their parametrization as Directed Acyclical Graphs (DAG)~\cite{pearl2009causality} representing the image ($X$), target label ($Y$), causal factor ($A$), and confounder ($C$). The multimodal model consists of three prediction branches: the main branch is for predicting MACE ($Y$), while the other two discriminative branches are predicting causal factors ($Y_{\text{ca}}$) and confounder labels ($Y_{\text{conf}}$) using features from the early transformer layers of the vision transformer encoder. 
\subsection{Single Modality Encoder}
For step 1, we design the de-confounding through distinct 12 layers ViT encoders for processing CXR and ECG images ($X_{CXR}$ \& $X_{ECG}$). The causal relationship $X \rightarrow Y$ represents the desired effect, however the backdoor path $X \leftarrow C \rightarrow Y$ highlights the influence of the confounder $C$, which introduces spurious correlations and affects predictions (Figure~\ref{fig:multimdal framework} a). To mitigate this bias, we remove the confounding influence separately from each encoder by eliminating the wrong links $C \to X$ and $C \to Y$ using two separate backpropagation graphs. The first loss function focuses on the main MACE predictions task: $\text{Loss}_{\text{Main}} = L_{\text{CE}}(y, \hat{y})$ and the second loss is dedicated to confounder task, handling gender and age predictions using cross-entropy: $\text{Loss}_{\text{Conf}} = L_{\text{CE}}(y_{\text{c}}, \hat{y}_{\text{c}}) + \alpha L_{\text{Conf}}(\hat{y}_{\text{c}})$. To further mitigate confounder influence, we incorporate confusion loss $\alpha L_{\text{Conf}}(\hat{y}_{\text{c}}$), encouraging the confounder predictions to approximate a uniform distribution~\cite{DINSDALE2021117689}. To effectively mitigate confounders, we utilize features from the early layers of the ViT which primarily capture low-level patterns and naturally encode strong confounding signals related to factors like gender and age. 


\subsection{Multimodal Co-attention alignment}
We utilize the pre-trained encoder weights from Step 1, freezing them in the subsequent co-attention module (Fig.~\ref{fig:multimdal framework} d). However, the primary challenge for the ECG and CXR alignment is unknown spatial relation. To preserve the primary knowledge from CXR and use ECG as a guiding modality, we adapted the co-attention mechanism in H\&E slides processing with genomic factor~\cite{9710773}. ECG features (\(E\)) serve as the query, while CXR features (\(CXR\)) function as both the key and value in the co-attention computation: $\text{CoAttn}_{\text{E} \rightarrow \text{CXR}} (E, C) = \text{softmax} \left( \frac{Q_{CXR} K_E^\top}{\sqrt{d_k}} \right) V_E$. The attention scores are obtained by computing the similarity between the projected query (\(Q = W_q^E E\)) and key (\(K = W_k^{CXR} CXR\)), scaled by \( d_k \) for numerical stability. A softmax operation is applied to produce the co-attention matrix (\( A_{\text{coattn}} \in \mathbb{R}^{N \times M} \)), which assigns importance weights to different regions of \(CXR\). These attention weights are used to compute a weighted sum of the CXR features, refining them into \( \hat{CXR} \) as: 

$\text{softmax} \left( \frac{W_q^E E {CXR}^\top W_k^{CXR}}{\sqrt{d_k}} \right) W_v^{CXR} {CXR}\rightarrow A_{\text{coattn}} W_v^{CXR} {CXR} \rightarrow \hat{CXR}$ 
,where \( W_q^E, W_k^{CXR}, W_v^{CXR} \in \mathbb{R}^{d_k \times d_k} \) are trainable weights. The values are derived from the CXR features as \( V = W_v^{CXR} CXR \) which ensures that the model selectively attends to the most informative regions, conditioned on ECG features. 



After the co-attention mechanism, the CXR and ECG features are processed through two transformers and multi-layer perceptron (MLP) branch to perform the classification tasks. The model consists of five branches, addressing two distinct tasks: the causal task, which predicts the classification label \( Y_{\text{ca}} \), and the main task, which predicts the classification label \( Y \). As shown in Fig.~\ref{fig:multimdal framework} b the relationship $X \leftarrow A \rightarrow Y$ indicates that causal factors $A$ influence both $X$ and $Y$. Our goal is to allow the model to leverage information from $A$ and improve generalization in MACE prediction. Causal learning is applied separately to both CXR and ECG features to preserve the heterogeneity of modality-specific information and their combined representations for improved classification performance. The loss function of five branches classification task is presented as: $\mathcal{L}_{\text{total}} = \mathcal{L}_{\text{causal}}^{\text{CXR}} + \mathcal{L}_{\text{main}}^{\text{CXR}} + \mathcal{L}_{\text{causal}}^{\text{ECG}} + \mathcal{L}_{\text{main}}^{\text{ECG}} + \mathcal{L}_{\text{main}}^{\text{concat}}$, where $\mathcal{L}_{\text{causal}}^{\text{CXR}}$ and $\mathcal{L}_{\text{main}}^{\text{CXR}}$ denote the causal and main losses for CXR, $\mathcal{L}_{\text{causal}}^{\text{ECG}}$ and $\mathcal{L}_{\text{main}}^{\text{ECG}}$ represent the causal and main losses for ECG, and $\mathcal{L}_{\text{main}}^{\text{concat}}$ is the combined loss for concatenated ECG and CXR.



\section{Results}


\textbf{Internal cohort.} Upon approval Institutional Review Board, retrospective data for 21,872 patients were collected across geographically disparate four sites. Patients who were lost to follow-up within 6 months were excluded due to uncertainty regarding the outcomes. Subsequent MACE outcomes were compiled by manual chart-review. Ultimately, the cohort included 12,612 patients who CXR performed and had a matching 12-lead ECG within a ±6 months time window. The dataset was randomly split at the patient level (80\% training, 20\% testing), yielding 79,648 and 21,108 image pairs, respectively. The dataset consists of high-risk cardiac patients, with elevated 51.77\% and 55.91\% MACE event rate within 6 month and one year respectively.

\noindent\textbf{Shift cohort (ED).} To evaluate the model's generalization  across `shift' data, we collected testing data from patient with CXR for any indication performed at Emergency Department (ED). The Shift cohort consisted of 2,638 patients and ECG signals were matched with CXR yielding 2,830 image-ECG pair. Compared to training population, this cohort includes younger patients (8.34\% over 80yrs) with significantly low presence of comorbidities, such as CHF (0\%), CKD (0\%), diabetes (0.94\%) and hypertension (2\%), as well as low MACE rate (25.12\%) within one year. 

\noindent\textbf{Shift cohort (MIMIC).} To benchmark, we evaluated the model on publicly available MIMICIV eICU data~\cite{johnson2020mimic}. We randomly selected 200 patients who underwent inpatient CXR imaging for any clinical indication. Among them, we identified 175 patients with temporally matched ECG recordings. We curated the 6 month MACE outcome by parsing the clinic notes and inpatient mortality data and MACE rate was 25.71\%. 



\noindent \textbf{Performance}: We compared single modality and baseline multimodal models against our proposed confounder debiasing (Conf), causal+confounder (With Confounder), and causal only (Without Confounder) models. The baseline single modality encoders are derived from MedCLIP and are fine-tuned separately for CXR and ECG data(\emph{CXR} and \emph{ECG} model respectively). We compared the proposed multimodal alignment framework MOSCARD against - MedCLIP (contrastive) and ALBEF (knowledge distillation) baselines~\cite{wang2022medclip,li2021align}. The evaluation results for single and multimodality models are summarized in Table~\ref{table:single_modality} and~\ref{table:multi_modality}.

On the internal datasets, \emph{Baseline CXR} achieves the best performance, particularly for MACE prediction at 5 years, with an AUC of 0.739 (Table~\ref{table:single_modality}). However, incorporating the de-confounding branch performance does not significantly improve on the internal testset, indicating that de-confounding does not enhance predictive performance on the hold-out test; however, the performance is on the par with the baseline (CXR Conf 5 years AUC 0.722). De-confounding models demonstrate optimal generalizability on external datasets (ED: 0.778 CXR Conf and MIMIC: 0.66 AUC ECG Conf), emphasizing that baseline models without de-confounding may rely on shortcut features for prediction, which are not generalizable. 
\begin{table}[tb!]
\centering
\caption{\footnotesize Comparison of single modality models with and without the de-confounding (95\% confidence intervals using bootstrapping). Green cells highlight the optimal performance.}
\resizebox{0.75\textwidth}{!}{%
\begin{tabular}{|ll|ll|ll|ll|ll|}
\hline
\multicolumn{2}{|l|}{\textbf{Single Modality}} & \multicolumn{2}{l|}{\textbf{CXR}} & \multicolumn{2}{l|}{\textbf{ECG}} & \multicolumn{2}{l|}{\textbf{CXR Conf}} & \multicolumn{2}{l|}{\textbf{ECG Conf}} \\ \hline
\multicolumn{1}{|l|}{} & \textit{\textbf{MACE}} & \multicolumn{1}{l|}{\textit{\textbf{Accuracy}}} & \textit{\textbf{AUC}} & \multicolumn{1}{l|}{\textit{\textbf{Accuracy}}} & \textit{\textbf{AUC}} & \multicolumn{1}{l|}{\textit{\textbf{Accuracy}}} & \textit{\textbf{AUC}} & \multicolumn{1}{l|}{\textit{\textbf{Accuracy}}} & \textit{\textbf{AUC}} \\ \cline{2-10} 
\multicolumn{1}{|l|}{} & 6M & \multicolumn{1}{l|}{\cellcolor[HTML]{67FD9A}\begin{tabular}[c]{@{}l@{}}0.656\\ {[}0.653, 0.659{]}\end{tabular}} & \cellcolor[HTML]{67FD9A}\begin{tabular}[c]{@{}l@{}}0.710\\ {[}0.707, 0.715{]}\end{tabular} & \multicolumn{1}{l|}{\begin{tabular}[c]{@{}l@{}}0.594\\ {[}0.593, 0.598{]}\end{tabular}} & \begin{tabular}[c]{@{}l@{}}0.633\\ {[}0.625, 0.634{]}\end{tabular} & \multicolumn{1}{l|}{\begin{tabular}[c]{@{}l@{}}0.639\\ {[}0.639, 0.646{]}\end{tabular}} & \begin{tabular}[c]{@{}l@{}}0.694\\ {[}0.693, 0.700{]}\end{tabular} & \multicolumn{1}{l|}{\begin{tabular}[c]{@{}l@{}}0.607\\ {[}0.604, 0.612{]}\end{tabular}} & \begin{tabular}[c]{@{}l@{}}0.648\\ {[}0.646, 0.654{]}\end{tabular} \\ \cline{2-10} 
\multicolumn{1}{|l|}{} & 1yr & \multicolumn{1}{l|}{\cellcolor[HTML]{67FD9A}\begin{tabular}[c]{@{}l@{}}0.664\\ {[}0.660, 0.668{]}\end{tabular}} & \cellcolor[HTML]{67FD9A}\begin{tabular}[c]{@{}l@{}}0.715\\ {[}0.711, 0.717{]}\end{tabular} & \multicolumn{1}{l|}{\begin{tabular}[c]{@{}l@{}}0.599\\ {[}0.594, 0.601{]}\end{tabular}} & \begin{tabular}[c]{@{}l@{}}0.640\\ {[}0.637, 0.647{]}\end{tabular} & \multicolumn{1}{l|}{\begin{tabular}[c]{@{}l@{}}0.646\\ {[}0.640, 0.647{]}\end{tabular}} & \begin{tabular}[c]{@{}l@{}}0.698\\ {[}0.693, 0.701{]}\end{tabular} & \multicolumn{1}{l|}{\begin{tabular}[c]{@{}l@{}}0.616\\ {[}0.611, 0.618{]}\end{tabular}} & \begin{tabular}[c]{@{}l@{}}0.656\\ {[}0.656, 0.661{]}\end{tabular} \\ \cline{2-10} 
\multicolumn{1}{|l|}{} & 2yr & \multicolumn{1}{l|}{\cellcolor[HTML]{67FD9A}\begin{tabular}[c]{@{}l@{}}0.665\\ {[}0.666, 0.673{]}\end{tabular}} & \cellcolor[HTML]{67FD9A}\begin{tabular}[c]{@{}l@{}}0.719\\ {[}0.713, 0.724{]}\end{tabular} & \multicolumn{1}{l|}{\begin{tabular}[c]{@{}l@{}}0.605\\ {[}0.603, 0.611{]}\end{tabular}} & \begin{tabular}[c]{@{}l@{}}0.644\\ {[}0.642, 0.647{]}\end{tabular} & \multicolumn{1}{l|}{\begin{tabular}[c]{@{}l@{}}0.651\\ {[}0.649, 0.656{]}\end{tabular}} & \begin{tabular}[c]{@{}l@{}}0.702\\ {[}0.699, 0.710{]}\end{tabular} & \multicolumn{1}{l|}{\begin{tabular}[c]{@{}l@{}}0.618\\ {[}0.612, 0.620{]}\end{tabular}} & \begin{tabular}[c]{@{}l@{}}0.657\\ {[}0.655, 0.661{]}\end{tabular} \\ \cline{2-10} 
\multicolumn{1}{|l|}{\multirow{-5}{*}{\textit{Internal}}} & 5yr & \multicolumn{1}{l|}{\cellcolor[HTML]{67FD9A}\begin{tabular}[c]{@{}l@{}}0.685\\ {[}0.679, 0.684{]}\end{tabular}} & \cellcolor[HTML]{67FD9A}\begin{tabular}[c]{@{}l@{}}0.739\\ {[}0.735, 0.745{]}\end{tabular} & \multicolumn{1}{l|}{\begin{tabular}[c]{@{}l@{}}0.603\\ {[}0.602, 0.607{]}\end{tabular}} & \begin{tabular}[c]{@{}l@{}}0.647\\ {[}0.645, 0.652{]}\end{tabular} & \multicolumn{1}{l|}{\begin{tabular}[c]{@{}l@{}}0.668\\ {[}0.665, 0.670{]}\end{tabular}} & \begin{tabular}[c]{@{}l@{}}0.722\\ {[}0.715, 0.725{]}\end{tabular} & \multicolumn{1}{l|}{\begin{tabular}[c]{@{}l@{}}0.614\\ {[}0.611, 0.617{]}\end{tabular}} & \begin{tabular}[c]{@{}l@{}}0.661\\ {[}0.660, 0.669{]}\end{tabular} \\ \hline
\multicolumn{1}{|l|}{\textit{\begin{tabular}[c]{@{}l@{}}External \\ (MIMIC)\end{tabular}}} & 6M & \multicolumn{1}{l|}{\begin{tabular}[c]{@{}l@{}}0.560\\ {[}0.546, 0.575{]}\end{tabular}} & \begin{tabular}[c]{@{}l@{}}0.642\\ {[}0.630, 0.658{]}\end{tabular} & \multicolumn{1}{l|}{\begin{tabular}[c]{@{}l@{}}0.663\\ {[}0.663, 0.687{]}\end{tabular}} & \begin{tabular}[c]{@{}l@{}}0.699\\ {[}0.694, 0.725{]}\end{tabular} & \multicolumn{1}{l|}{\begin{tabular}[c]{@{}l@{}}0.560\\ {[}0.547, 0.568{]}\end{tabular}} & \begin{tabular}[c]{@{}l@{}}0.612\\ {[}0.600, 0.627{]}\end{tabular} & \multicolumn{1}{l|}{\cellcolor[HTML]{67FD9A}\begin{tabular}[c]{@{}l@{}}0.651\\ {[}0.645, 0.668{]}\end{tabular}} & \cellcolor[HTML]{67FD9A}\begin{tabular}[c]{@{}l@{}}0.660\\ {[}0.645, 0.679{]}\end{tabular} \\ \hline
\multicolumn{1}{|l|}{\textit{\begin{tabular}[c]{@{}l@{}}External \\ (ED)\end{tabular}}} & 1yr & \multicolumn{1}{l|}{\begin{tabular}[c]{@{}l@{}}0.724\\ {[}0.720, 0.727{]}\end{tabular}} & \begin{tabular}[c]{@{}l@{}}0.768\\ {[}0.764, 0.777{]}\end{tabular} & \multicolumn{1}{l|}{\begin{tabular}[c]{@{}l@{}}0.673\\ {[}0.669, 0.676{]}\end{tabular}} & \begin{tabular}[c]{@{}l@{}}0.742\\ {[}0.733, 0.749{]}\end{tabular} & \multicolumn{1}{l|}{\cellcolor[HTML]{67FD9A}\begin{tabular}[c]{@{}l@{}}0.712\\ {[}0.710, 0.716{]}\end{tabular}} & \cellcolor[HTML]{67FD9A}\begin{tabular}[c]{@{}l@{}}0.778\\ {[}0.774, 0.784{]}\end{tabular} & \multicolumn{1}{l|}{\begin{tabular}[c]{@{}l@{}}0.670\\ {[}0.666, 0.676{]}\end{tabular}} & \begin{tabular}[c]{@{}l@{}}0.724\\ {[}0.712, 0.726{]}\end{tabular} \\ \hline
\end{tabular}%
}
\label{table:single_modality}
\end{table}

\begin{table}[tb!]
\caption{\footnotesize Comparative model performance after multimodal data alignment with combined and individual modality (95\% confidence intervals using bootstrapping). Green cells highlights the optimal performance. }
\label{table:multi_modality}
\centering
\resizebox{0.68\textwidth}{!}{%
\begin{tabular}{|llllllll|}
\hline
\multicolumn{8}{|c|}{{\color[HTML]{000000} \textbf{Knowledge distillation Baseline (ALBEF)}}} \\ \hline
\multicolumn{1}{|l|}{{\color[HTML]{000000} \textbf{}}} & \multicolumn{1}{l|}{{\color[HTML]{000000} \textbf{MACE}}} & \multicolumn{2}{l|}{{\color[HTML]{000000} \textbf{CXR+ECG (Combined)}}} & \multicolumn{2}{l|}{{\color[HTML]{000000} \textbf{CXR}}} & \multicolumn{2}{l|}{{\color[HTML]{000000} \textbf{ECG}}} \\ \hline
\multicolumn{1}{|l|}{{\color[HTML]{000000} }} & \multicolumn{1}{l|}{{\color[HTML]{000000} \textit{\textbf{Mace}}}} & \multicolumn{1}{l|}{{\color[HTML]{000000} \textit{\textbf{Accuracy}}}} & \multicolumn{1}{l|}{{\color[HTML]{000000}\textit{\textbf{AUC}} }} & \multicolumn{1}{l|}{{\color[HTML]{000000} \textit{\textbf{Accuracy}}}} & \multicolumn{1}{l|}{{\color[HTML]{000000} \textit{\textbf{AUC}}}} & \multicolumn{1}{l|}{{\color[HTML]{000000}\textit{\textbf{Accuracy}} }} & {\color[HTML]{000000} \textit{\textbf{AUC}}} \\ \cline{2-8} 
\multicolumn{1}{|l|}{{\color[HTML]{000000} }} & \multicolumn{1}{l|}{{\color[HTML]{000000} 6M}} & \multicolumn{1}{l|}{{\color[HTML]{000000}\begin{tabular}[c]{@{}l@{}}0.571\\ {[}0.535,0.635{]}\end{tabular} }} & \multicolumn{1}{l|}{{\color[HTML]{000000}\begin{tabular}[c]{@{}l@{}}0.608\\ {[}0.531,0.648{]}\end{tabular} }} & \multicolumn{1}{l|}{{\color[HTML]{000000}\begin{tabular}[c]{@{}l@{}}0.631\\ {[}0.588,0.688{]}\end{tabular} }} & \multicolumn{1}{l|}{{\color[HTML]{000000}\begin{tabular}[c]{@{}l@{}}0.512\\ {[}0.467,0.560{]}\end{tabular} }} & \multicolumn{1}{l|}{{\color[HTML]{000000}\begin{tabular}[c]{@{}l@{}}0.568\\ {[}0.537,0.615{]}\end{tabular} }} & {\color[HTML]{000000}\begin{tabular}[c]{@{}l@{}}0.600\\ {[}0.513,0.615{]}\end{tabular} } \\ \cline{2-8} 
\multicolumn{1}{|l|}{{\color[HTML]{000000} }} & \multicolumn{1}{l|}{{\color[HTML]{000000} 1yr}} & \multicolumn{1}{l|}{{\color[HTML]{000000} \begin{tabular}[c]{@{}l@{}}0.595\\ {[}0.555,0.638{]}\end{tabular}}} & \multicolumn{1}{l|}{{\color[HTML]{000000}\begin{tabular}[c]{@{}l@{}}0.612\\ {[}0.525,0.675{]}\end{tabular} }} & \multicolumn{1}{l|}{{\color[HTML]{000000} \begin{tabular}[c]{@{}l@{}}0.572\\ {[}0.532,0.613{]}\end{tabular}}} & \multicolumn{1}{l|}{{\color[HTML]{000000}\begin{tabular}[c]{@{}l@{}}0.549\\ {[}0.494,0.592{]}\end{tabular}}} & \multicolumn{1}{l|}{{\color[HTML]{000000}\begin{tabular}[c]{@{}l@{}}0.580\\ {[}0.543,0.633{]}\end{tabular}  }} & {\color[HTML]{000000}\begin{tabular}[c]{@{}l@{}}0.626\\ {[}0.528,0.640{]}\end{tabular}  } \\ \cline{2-8} 
\multicolumn{1}{|l|}{{\color[HTML]{000000} }} & \multicolumn{1}{l|}{{\color[HTML]{000000} 2yr}} & \multicolumn{1}{l|}{{\color[HTML]{000000}\begin{tabular}[c]{@{}l@{}}0.562\\ {[}0.511,0.618{]}\end{tabular} }} & \multicolumn{1}{l|}{{\color[HTML]{000000} \begin{tabular}[c]{@{}l@{}}0.587\\ {[}0.512,0.619{]}\end{tabular}}} & \multicolumn{1}{l|}{{\color[HTML]{000000}\begin{tabular}[c]{@{}l@{}}0.507\\ {[}0.459,0.557{]}\end{tabular} }} & \multicolumn{1}{l|}{{\color[HTML]{000000} \begin{tabular}[c]{@{}l@{}}0.506\\ {[}0.457,0.562{]}\end{tabular}}} & \multicolumn{1}{l|}{{\color[HTML]{000000}\begin{tabular}[c]{@{}l@{}}0.570\\ {[}0.534,0.625{]}\end{tabular} }} & {\color[HTML]{000000}\begin{tabular}[c]{@{}l@{}}0.626\\ {[}0.534,0.630{]}\end{tabular} } \\ \cline{2-8} 
\multicolumn{1}{|l|}{\multirow{-5}{*}{{\color[HTML]{000000} \textit{Internal}}}} & \multicolumn{1}{l|}{{\color[HTML]{000000} 5yr}} & \multicolumn{1}{l|}{{\color[HTML]{000000}\begin{tabular}[c]{@{}l@{}}0.596\\ {[}0.554,0.645{]}\end{tabular} }} & \multicolumn{1}{l|}{{\color[HTML]{000000}\begin{tabular}[c]{@{}l@{}}0.617\\ {[}0.555,0.643{]}\end{tabular} }} & \multicolumn{1}{l|}{{\color[HTML]{000000} \begin{tabular}[c]{@{}l@{}}0.509\\ {[}0.477,0.556{]}\end{tabular}}} & \multicolumn{1}{l|}{{\color[HTML]{000000}\begin{tabular}[c]{@{}l@{}}0.510\\ {[}0.473,0.563{]}\end{tabular} 
}} & \multicolumn{1}{l|}{{\color[HTML]{000000}\begin{tabular}[c]{@{}l@{}}0.579\\ {[}0.544,0.625{]}\end{tabular} }} & {\color[HTML]{000000}\begin{tabular}[c]{@{}l@{}}0.628\\ {[}0.541,0.630{]}\end{tabular} } \\ \hline
\multicolumn{1}{|l|}{{\color[HTML]{000000} \textit{\begin{tabular}[c]{@{}l@{}}External\\ (MIMIC)\end{tabular}}}} & \multicolumn{1}{l|}{{\color[HTML]{000000} 6M}} & \multicolumn{1}{l|}{{\color[HTML]{000000}\begin{tabular}[c]{@{}l@{}}0.531\\ {[}0.368,0.617{]}\end{tabular} }} & \multicolumn{1}{l|}{{\color[HTML]{000000}\begin{tabular}[c]{@{}l@{}}0.562\\ {[}0.349,0.640{]}\end{tabular} }} & \multicolumn{1}{l|}{{\color[HTML]{000000}\begin{tabular}[c]{@{}l@{}}0.668\\ {[}0.571,0.764{]}\end{tabular} }} & \multicolumn{1}{l|}{{\color[HTML]{000000}\begin{tabular}[c]{@{}l@{}}0.542\\ {[}0.430,0.663{]}\end{tabular} }} & \multicolumn{1}{l|}{{\color[HTML]{000000} \begin{tabular}[c]{@{}l@{}}0.588\\ {[}0.444,0.675{]}\end{tabular} }} & {\color[HTML]{000000}\begin{tabular}[c]{@{}l@{}}0.604\\ {[}0.422,0.694{]}\end{tabular}  } \\ \hline
\multicolumn{1}{|l|}{{\color[HTML]{000000} \textit{\begin{tabular}[c]{@{}l@{}}External \\ (ED)\end{tabular}}}} & \multicolumn{1}{l|}{{\color[HTML]{000000} 1yr}} & \multicolumn{1}{l|}{{\color[HTML]{000000}\begin{tabular}[c]{@{}l@{}}0.589\\ {[}0.557, 0.649{]}\end{tabular} }} & \multicolumn{1}{l|}{{\color[HTML]{000000}\begin{tabular}[c]{@{}l@{}}0.603\\ {[}0.498, 0.666{]}\end{tabular} }} & \multicolumn{1}{l|}{{\color[HTML]{000000} \begin{tabular}[c]{@{}l@{}}0.523\\ {[}0.486,0.548{]}\end{tabular}}} & \multicolumn{1}{l|}{{\color[HTML]{000000} \begin{tabular}[c]{@{}l@{}}0.508\\ {[}0.449,0.589{]}\end{tabular}}} & \multicolumn{1}{l|}{{\color[HTML]{000000} \begin{tabular}[c]{@{}l@{}}0.669\\ {[}0.649,0.690{]}\end{tabular}}} & {\color[HTML]{000000}\begin{tabular}[c]{@{}l@{}}0.732\\ {[}0.647,0.695{]}\end{tabular} } \\ \hline
\multicolumn{8}{|c|}{{\color[HTML]{000000} \textbf{Contrastive Alignment Baseline (MedCLIP)}}} \\ \hline
\multicolumn{1}{|l|}{{\color[HTML]{000000} }} & \multicolumn{1}{l|}{{\color[HTML]{000000} }} & \multicolumn{2}{l|}{{\color[HTML]{000000} \textbf{CXR+ECG (Combined)}}} & \multicolumn{2}{l|}{{\color[HTML]{000000} \textbf{CXR+ECG (CXR)}}} & \multicolumn{2}{l|}{{\color[HTML]{000000} \textbf{CXR+ECG (ECG)}}} \\ \hline
\multicolumn{1}{|l|}{{\color[HTML]{000000} }} & \multicolumn{1}{l|}{{\color[HTML]{000000} \textit{\textbf{Mace}}}} & \multicolumn{1}{l|}{{\color[HTML]{000000} \textit{\textbf{Accuracy}}}} & \multicolumn{1}{l|}{{\color[HTML]{000000} \textit{\textbf{AUC}}}} & \multicolumn{1}{l|}{{\color[HTML]{000000} \textit{\textbf{Accuracy}}}} & \multicolumn{1}{l|}{{\color[HTML]{000000} \textit{\textbf{AUC}}}} & \multicolumn{1}{l|}{{\color[HTML]{000000} \textit{\textbf{Accuracy}}}} & {\color[HTML]{000000} \textit{\textbf{AUC}}} \\ \cline{2-8} 
\multicolumn{1}{|l|}{{\color[HTML]{000000} }} & \multicolumn{1}{l|}{{\color[HTML]{000000} 6M}} & \multicolumn{1}{l|}{{\color[HTML]{000000} \begin{tabular}[c]{@{}l@{}}0.585\\ {[}0.583,0.586{]}\end{tabular}}} & \multicolumn{1}{l|}{{\color[HTML]{000000} \begin{tabular}[c]{@{}l@{}}0.613\\ {[}0.610, 616{]}\end{tabular}}} & \multicolumn{1}{l|}{{\color[HTML]{000000} \begin{tabular}[c]{@{}l@{}}0.486\\ {[}0.484, 0.487{]}\end{tabular}}} & \multicolumn{1}{l|}{{\color[HTML]{000000} \begin{tabular}[c]{@{}l@{}}0.482\\ {[}0.480, 0.485{]}\end{tabular}}} & \multicolumn{1}{l|}{{\color[HTML]{000000} \begin{tabular}[c]{@{}l@{}}0.498\\ {[}0.497, 0.500{]}\end{tabular}}} & {\color[HTML]{000000} \begin{tabular}[c]{@{}l@{}}0.496\\ {[}0.494, 0.503{]}\end{tabular}} \\ \cline{2-8} 
\multicolumn{1}{|l|}{{\color[HTML]{000000} }} & \multicolumn{1}{l|}{{\color[HTML]{000000} 1yr}} & \multicolumn{1}{l|}{{\color[HTML]{000000} \begin{tabular}[c]{@{}l@{}}0.593\\ {[}0.593, 0.595{]}\end{tabular}}} & \multicolumn{1}{l|}{{\color[HTML]{000000} \begin{tabular}[c]{@{}l@{}}0.628\\ {[}0.626, 0.635{]}\end{tabular}}} & \multicolumn{1}{l|}{{\color[HTML]{000000} \begin{tabular}[c]{@{}l@{}}0.494\\ {[}0.490, 0.493{]}\end{tabular}}} & \multicolumn{1}{l|}{{\color[HTML]{000000} \begin{tabular}[c]{@{}l@{}}0.492\\ {[}0.491, 0.495{]}\end{tabular}}} & \multicolumn{1}{l|}{{\color[HTML]{000000} \begin{tabular}[c]{@{}l@{}}0.500\\ {[}0.496, 0.501{]}\end{tabular}}} & {\color[HTML]{000000} \begin{tabular}[c]{@{}l@{}}0.499\\ {[}0.496, 0.499{]}\end{tabular}} \\ \cline{2-8} 
\multicolumn{1}{|l|}{{\color[HTML]{000000} }} & \multicolumn{1}{l|}{{\color[HTML]{000000} 2yr}} & \multicolumn{1}{l|}{{\color[HTML]{000000} \begin{tabular}[c]{@{}l@{}}0.583\\ {[}0.581,0.585{]}\end{tabular}}} & \multicolumn{1}{l|}{{\color[HTML]{000000} \begin{tabular}[c]{@{}l@{}}0.613\\ {[}0.610, 613{]}\end{tabular}}} & \multicolumn{1}{l|}{{\color[HTML]{000000} \begin{tabular}[c]{@{}l@{}}0.486\\ {[}0.485, 0.488{]}\end{tabular}}} & \multicolumn{1}{l|}{{\color[HTML]{000000} \begin{tabular}[c]{@{}l@{}}0.477\\ {[}0.475, 0.479{]}\end{tabular}}} & \multicolumn{1}{l|}{{\color[HTML]{000000} \begin{tabular}[c]{@{}l@{}}0.464\\ {[}0.463, 0.467{]}\end{tabular}}} & {\color[HTML]{000000} \begin{tabular}[c]{@{}l@{}}0.452\\ {[}0.450, 0.454{]}\end{tabular}} \\ \cline{2-8} 
\multicolumn{1}{|l|}{\multirow{-5}{*}{{\color[HTML]{000000} \textit{Internal}}}} & \multicolumn{1}{l|}{{\color[HTML]{000000} 5yr}} & \multicolumn{1}{l|}{{\color[HTML]{000000} \begin{tabular}[c]{@{}l@{}}0.573\\ {[}0.570, 0.573{]}\end{tabular}}} & \multicolumn{1}{l|}{{\color[HTML]{000000} \begin{tabular}[c]{@{}l@{}}0.601\\ {[}0.597, 0.601{]}\end{tabular}}} & \multicolumn{1}{l|}{{\color[HTML]{000000} \begin{tabular}[c]{@{}l@{}}0.496\\ {[}0.495, 0.498{]}\end{tabular}}} & \multicolumn{1}{l|}{{\color[HTML]{000000} \begin{tabular}[c]{@{}l@{}}0.496\\ {[}0.495, 0.500{]}\end{tabular}}} & \multicolumn{1}{l|}{{\color[HTML]{000000} \begin{tabular}[c]{@{}l@{}}0.466\\ {[}0.463, 0.468{]}\end{tabular}}} & {\color[HTML]{000000} \begin{tabular}[c]{@{}l@{}}0.457\\ {[}0.456, 0.458{]}\end{tabular}} \\ \hline
\multicolumn{1}{|l|}{{\color[HTML]{000000} \textit{\begin{tabular}[c]{@{}l@{}}External\\ (MIMIC)\end{tabular}}}} & \multicolumn{1}{l|}{{\color[HTML]{000000} 6M}} & \multicolumn{1}{l|}{{\color[HTML]{000000} \begin{tabular}[c]{@{}l@{}}0.509\\ {[}0.485, 0.511{]}\end{tabular}}} & \multicolumn{1}{l|}{{\color[HTML]{000000} \begin{tabular}[c]{@{}l@{}}0.501\\ {[}0.484, 0.519{]}\end{tabular}}} & \multicolumn{1}{l|}{{\color[HTML]{000000} \begin{tabular}[c]{@{}l@{}}0.571\\ {[}0.559, 0.584{]}\end{tabular}}} & \multicolumn{1}{l|}{{\color[HTML]{000000} \begin{tabular}[c]{@{}l@{}}0.550\\ {[}0.530, 0.560{]}\end{tabular}}} & \multicolumn{1}{l|}{{\color[HTML]{000000} \begin{tabular}[c]{@{}l@{}}0.474\\ {[}0.460, 0.484{]}\end{tabular}}} & {\color[HTML]{000000} \begin{tabular}[c]{@{}l@{}}0.467\\ {[}0.448, 0.484{]}\end{tabular}} \\ \hline
\multicolumn{1}{|l|}{{\color[HTML]{000000} \textit{\begin{tabular}[c]{@{}l@{}}External \\ (ED)\end{tabular}}}} & \multicolumn{1}{l|}{{\color[HTML]{000000} 1yr}} & \multicolumn{1}{l|}{{\color[HTML]{000000} \begin{tabular}[c]{@{}l@{}}0.522\\ {[}0.520, 0.531{]}\end{tabular}}} & \multicolumn{1}{l|}{{\color[HTML]{000000} \begin{tabular}[c]{@{}l@{}}0.488\\ {[}0.484, 0.497{]}\end{tabular}}} & \multicolumn{1}{l|}{{\color[HTML]{000000} \begin{tabular}[c]{@{}l@{}}0.517\\ {[}0.511, 0.519{]}\end{tabular}}} & \multicolumn{1}{l|}{{\color[HTML]{000000} \begin{tabular}[c]{@{}l@{}}0.528\\ {[}0.521, 0.530{]}\end{tabular}}} & \multicolumn{1}{l|}{{\color[HTML]{000000} \begin{tabular}[c]{@{}l@{}}0.489\\ {[}0.483, 0.492{]}\end{tabular}}} & {\color[HTML]{000000} \begin{tabular}[c]{@{}l@{}}0.487\\ {[}0.484, 0.495{]}\end{tabular}} \\ \hline
\multicolumn{7}{|c|}{{\color[HTML]{000000} \textbf{MOSCARD}}} & {\color[HTML]{000000} } \\ \hline
\multicolumn{2}{|c|}{{\color[HTML]{000000} \textbf{With Confounder}}} & \multicolumn{2}{c|}{{\color[HTML]{000000} \textbf{CXR+ECG (Combined)}}} & \multicolumn{2}{c|}{{\color[HTML]{000000} \textbf{CXR+ECG (CXR)}}} & \multicolumn{2}{c|}{{\color[HTML]{000000} \textbf{CXR+ECG (ECG)}}} \\ \hline
\multicolumn{1}{|l|}{{\color[HTML]{000000} }} & \multicolumn{1}{l|}{{\color[HTML]{000000} \textit{\textbf{Mace}}}} & \multicolumn{1}{l|}{{\color[HTML]{000000} \textit{\textbf{Accuracy}}}} & \multicolumn{1}{l|}{{\color[HTML]{000000} \textit{\textbf{AUC}}}} & \multicolumn{1}{l|}{{\color[HTML]{000000} \textit{\textbf{Accuracy}}}} & \multicolumn{1}{l|}{{\color[HTML]{000000} \textit{\textbf{AUC}}}} & \multicolumn{1}{l|}{{\color[HTML]{000000} \textit{\textbf{Accuracy}}}} & {\color[HTML]{000000} \textit{\textbf{AUC}}} \\ \cline{2-8} 
\multicolumn{1}{|l|}{{\color[HTML]{000000} }} & \multicolumn{1}{l|}{{\color[HTML]{000000} 6M}} & \multicolumn{1}{l|}{{\cellcolor[HTML]{67FD9A}\color[HTML]{000000} \begin{tabular}[c]{@{}l@{}}0.681\\ {[}0.679, 0.682{]}\end{tabular}}} & \multicolumn{1}{l|}{{\cellcolor[HTML]{67FD9A}\color[HTML]{000000} \begin{tabular}[c]{@{}l@{}}0.733\\ {[}0.732, 0.735{]}\end{tabular}}} & \multicolumn{1}{l|}{{\color[HTML]{000000} \begin{tabular}[c]{@{}l@{}}0.670\\ {[}0.668, 0.670{]}\end{tabular}}} & \multicolumn{1}{l|}{{\color[HTML]{000000} \begin{tabular}[c]{@{}l@{}}0.716\\ {[}0.715, 0.718{]}\end{tabular}}} & \multicolumn{1}{l|}{{\color[HTML]{000000} \begin{tabular}[c]{@{}l@{}}0.673\\ {[}0.671, 0.674{]}\end{tabular}}} & {\color[HTML]{000000} \begin{tabular}[c]{@{}l@{}}0.723\\ {[}0.722, 0.725{]}\end{tabular}} \\ \cline{2-8} 
\multicolumn{1}{|l|}{{\color[HTML]{000000} }} & \multicolumn{1}{l|}{{\color[HTML]{000000} 1yr}} & \multicolumn{1}{l|}{{\cellcolor[HTML]{67FD9A}\color[HTML]{000000} \begin{tabular}[c]{@{}l@{}} 0.691\\ {[}0.689, 0.692{]}\end{tabular}}} & \multicolumn{1}{l|}{{\cellcolor[HTML]{67FD9A}\color[HTML]{000000} \begin{tabular}[c]{@{}l@{}} 0.750\\ {[}0.749, 0.752{]}\end{tabular}}} & \multicolumn{1}{l|}{{\color[HTML]{000000} \begin{tabular}[c]{@{}l@{}} 0.681\\ {[}0.676, 0.683{]}\end{tabular}}} & \multicolumn{1}{l|}{{\color[HTML]{000000} \begin{tabular}[c]{@{}l@{}}0.734\\ {[}0.733, 0.736{]}\end{tabular}}} & \multicolumn{1}{l|}{{\color[HTML]{000000} \begin{tabular}[c]{@{}l@{}}0.674\\ {[}0.672, 0.676{]}\end{tabular}}} & {\color[HTML]{000000} \begin{tabular}[c]{@{}l@{}}0.735\\ {[}0.734, 0.738{]}\end{tabular}} \\ \cline{2-8} 
\multicolumn{1}{|l|}{{\color[HTML]{000000} }} & \multicolumn{1}{l|}{{\color[HTML]{000000} 2yr}} & \multicolumn{1}{l|}{{\cellcolor[HTML]{67FD9A}\color[HTML]{000000} \begin{tabular}[c]{@{}l@{}} 0.683\\ {[}0.681, 0.684{]}\end{tabular}}} & \multicolumn{1}{l|}{{\cellcolor[HTML]{67FD9A}\color[HTML]{000000} \begin{tabular}[c]{@{}l@{}} 0.740\\ {[}0.738, 0.742{]}\end{tabular}}} & \multicolumn{1}{l|}{{\color[HTML]{000000} \begin{tabular}[c]{@{}l@{}} 0.669\\ {[}0.665, 0.670{]}\end{tabular}}}& \multicolumn{1}{l|}{{\color[HTML]{000000} \begin{tabular}[c]{@{}l@{}}0.723\\ {[}0.723, 0.726{]}\end{tabular}}} & \multicolumn{1}{l|}{{\color[HTML]{000000} \begin{tabular}[c]{@{}l@{}}0.675\\ {[}0.673, 0.676{]}\end{tabular}}} & {\color[HTML]{000000} \begin{tabular}[c]{@{}l@{}}0.733\\ {[}0.731, 0.734{]}\end{tabular}} \\ \cline{2-8} 
\multicolumn{1}{|l|}{\multirow{-5}{*}{{\color[HTML]{000000} \textit{Internal}}}} & \multicolumn{1}{l|}{{\color[HTML]{000000} 5yr}} & \multicolumn{1}{l|}{{\cellcolor[HTML]{67FD9A}\color[HTML]{000000} \begin{tabular}[c]{@{}l@{}} 0.678\\ {[}0.677, 0.679{]}\end{tabular}}} & \multicolumn{1}{l|}{{\cellcolor[HTML]{67FD9A}\color[HTML]{000000} \begin{tabular}[c]{@{}l@{}} 0.735\\ {[}0.733, 0.738{]}\end{tabular}}} & \multicolumn{1}{l|}{{\color[HTML]{000000} \begin{tabular}[c]{@{}l@{}} 0.669\\ {[}0.667, 0.670{]}\end{tabular}}} & \multicolumn{1}{l|}{{\color[HTML]{000000} \begin{tabular}[c]{@{}l@{}}0.717\\ {[}0.716, 0.719{]}\end{tabular}}} & \multicolumn{1}{l|}{{\color[HTML]{000000} \begin{tabular}[c]{@{}l@{}}0.675\\ {[}0.673, 0.676{]}\end{tabular}}} & {\color[HTML]{000000} \begin{tabular}[c]{@{}l@{}}0.730\\ {[}0.729, 0.735{]}\end{tabular}} \\ \hline
\multicolumn{1}{|l|}{{\color[HTML]{000000} \textit{\begin{tabular}[c]{@{}l@{}}External\\ (MIMIC)\end{tabular}}}} & \multicolumn{1}{l|}{{\color[HTML]{000000} 6M}} & \multicolumn{1}{l|}{{\color[HTML]{000000} \begin{tabular}[c]{@{}l@{}}0.623\\ {[}0.614, 0.635{]}\end{tabular}}} & \multicolumn{1}{l|}{{\color[HTML]{000000} \begin{tabular}[c]{@{}l@{}}0.662\\ {[}0.649, 0.676{]}\end{tabular}}} & \multicolumn{1}{l|}{{\color[HTML]{000000} \begin{tabular}[c]{@{}l@{}}0.600\\ {[}0.590, 0.615{]}\end{tabular}}} & \multicolumn{1}{l|}{{\color[HTML]{000000} \begin{tabular}[c]{@{}l@{}}0.644\\ {[}0.624, 0.658{]}\end{tabular}}} & \multicolumn{1}{l|}{\cellcolor[HTML]{FFFFFF}{\color[HTML]{000000} \begin{tabular}[c]{@{}l@{}}0.611\\ {[}0.594, 0.619{]}\end{tabular}}} & \cellcolor[HTML]{FFFFFF}{\color[HTML]{000000} \begin{tabular}[c]{@{}l@{}}0.657\\ {[}0.633, 0.670{]}\end{tabular}} \\ \hline
\multicolumn{1}{|l|}{{\color[HTML]{000000} \textit{\begin{tabular}[c]{@{}l@{}}External \\ (ED)\end{tabular}}}} & \multicolumn{1}{l|}{{\color[HTML]{000000} 1yr}} & \multicolumn{1}{l|}{{\color[HTML]{000000} \begin{tabular}[c]{@{}l@{}}0.702\\ {[}0.697, 0.704{]}\end{tabular}}} & \multicolumn{1}{l|}{{\color[HTML]{000000} \begin{tabular}[c]{@{}l@{}}0.777\\ {[}0.768, 0.783{]}\end{tabular}}} & \multicolumn{1}{l|}{\cellcolor[HTML]{FFFFFF}{\color[HTML]{000000} \begin{tabular}[c]{@{}l@{}}0.732\\ {[}0.727, 0.734{]}\end{tabular}}} & \multicolumn{1}{l|}{\cellcolor[HTML]{FFFFFF}{\color[HTML]{000000} \begin{tabular}[c]{@{}l@{}}0.809\\ {[}0.799, 0.811{]}\end{tabular}}} & \multicolumn{1}{l|}{{\color[HTML]{000000} \begin{tabular}[c]{@{}l@{}}0.647\\ {[}0.645, 0.653{]}\end{tabular}}} & {\color[HTML]{000000} \begin{tabular}[c]{@{}l@{}}0.710\\ {[}0.705, 0.714{]}\end{tabular}} \\ \hline
\multicolumn{2}{|c|}{{\color[HTML]{000000} \textbf{Without Confounder}}} & \multicolumn{2}{c|}{{\color[HTML]{000000} \textbf{CXR+ECG (Combined)}}} & \multicolumn{2}{c|}{{\color[HTML]{000000} \textbf{CXR+ECG (CXR)}}} & \multicolumn{2}{c|}{{\color[HTML]{000000} \textbf{CXR+ECG (ECG)}}} \\ \hline
\multicolumn{1}{|l|}{{\color[HTML]{000000} }} & \multicolumn{1}{l|}{{\color[HTML]{000000} \textit{\textbf{Mace}}}} & \multicolumn{1}{l|}{{\color[HTML]{000000} \textit{\textbf{Accuracy}}}} & \multicolumn{1}{l|}{{\color[HTML]{000000} \textit{\textbf{AUC}}}} & \multicolumn{1}{l|}{{\color[HTML]{000000} \textit{\textbf{Accuracy}}}} & \multicolumn{1}{l|}{{\color[HTML]{000000} \textit{\textbf{AUC}}}} & \multicolumn{1}{l|}{{\color[HTML]{000000} \textit{\textbf{Accuracy}}}} & {\color[HTML]{000000} \textit{\textbf{AUC}}} \\ \cline{2-8}
\multicolumn{1}{|l|}{{\color[HTML]{000000} }} & \multicolumn{1}{l|}{{\color[HTML]{000000} 6M}} & \multicolumn{1}{l|}{{\color[HTML]{000000} \begin{tabular}[c]{@{}l@{}}0.652\\ {[}0.649, 0.653{]}\end{tabular}}} & \multicolumn{1}{l|}{{\color[HTML]{000000} \begin{tabular}[c]{@{}l@{}}0.711\\ {[}0.707, 0.711{]}\end{tabular}}} & \multicolumn{1}{l|}{{\color[HTML]{000000} \begin{tabular}[c]{@{}l@{}}0.644\\ {[}0.643, 0.647{]}\end{tabular}}} & \multicolumn{1}{l|}{{\color[HTML]{000000} \begin{tabular}[c]{@{}l@{}}0.695\\ {[}0.692, 0.696{]}\end{tabular}}} & \multicolumn{1}{l|}{{\color[HTML]{000000} \begin{tabular}[c]{@{}l@{}}0.641\\ {[}0.640, 0.645{]}\end{tabular}}} & {\color[HTML]{000000} \begin{tabular}[c]{@{}l@{}}0.686\\ {[}0.684, 0.687{]}\end{tabular}} \\ \cline{2-8} 
\multicolumn{1}{|l|}{{\color[HTML]{000000} }} & \multicolumn{1}{l|}{{\color[HTML]{000000} 1yr}} & \multicolumn{1}{l|}{{\color[HTML]{000000} \begin{tabular}[c]{@{}l@{}}0.662\\{[}0.661, 0.663{]}\end{tabular}}} & \multicolumn{1}{l|}{{\color[HTML]{000000} \begin{tabular}[c]{@{}l@{}}0.724\\ {[}0.723, 0.726{]}\end{tabular}}} & \multicolumn{1}{l|}{{\color[HTML]{000000} \begin{tabular}[c]{@{}l@{}}0.656\\ {[}0.653, 0.658{]}\end{tabular}}} & \multicolumn{1}{l|}{{\color[HTML]{000000} \begin{tabular}[c]{@{}l@{}}0.713\\ {[}0.711, 0.714{]}\end{tabular}}} & \multicolumn{1}{l|}{{\color[HTML]{000000} \begin{tabular}[c]{@{}l@{}}0.647\\ {[}0.645, 0.648{]}\end{tabular}}} & {\color[HTML]{000000} \begin{tabular}[c]{@{}l@{}}0.694\\ {[}0.693, 0.696{]}\end{tabular}} \\ \cline{2-8} 
\multicolumn{1}{|l|}{{\color[HTML]{000000} }} & \multicolumn{1}{l|}{{\color[HTML]{000000} 2yr}} & \multicolumn{1}{l|}{{\color[HTML]{000000} \begin{tabular}[c]{@{}l@{}}0.654\\ {[}0.653, 0.656{]}\end{tabular}}} & \multicolumn{1}{l|}{{\color[HTML]{000000} \begin{tabular}[c]{@{}l@{}}0.715\\ {[}0.714, 0.718{]}\end{tabular}}} & \multicolumn{1}{l|}{{\color[HTML]{000000} \begin{tabular}[c]{@{}l@{}}0.648\\ {[}0.646, 0.649{]}\end{tabular}}} & \multicolumn{1}{l|}{{\color[HTML]{000000} \begin{tabular}[c]{@{}l@{}}0.706\\ {[}0.704, 0.708{]}\end{tabular}}} & \multicolumn{1}{l|}{{\color[HTML]{000000} \begin{tabular}[c]{@{}l@{}}0.649\\ {[}0.649, 0.652{]}\end{tabular}}} & {\color[HTML]{000000} \begin{tabular}[c]{@{}l@{}}0.695\\ {[}0.695, 0.698{]}\end{tabular}} \\ \cline{2-8} 
\multicolumn{1}{|l|}{\multirow{-4}{*}{{\color[HTML]{000000} \textit{Internal}}}} & \multicolumn{1}{l|}{{\color[HTML]{000000} 5yr}} & \multicolumn{1}{l|}{{\color[HTML]{000000} \begin{tabular}[c]{@{}l@{}}0.650\\ {[}0.647, 0.651{]}\end{tabular}}} & \multicolumn{1}{l|}{{\color[HTML]{000000} \begin{tabular}[c]{@{}l@{}}0.711\\ {[}0.708, 0.711{]}\end{tabular}}} & \multicolumn{1}{l|}{{\color[HTML]{000000} \begin{tabular}[c]{@{}l@{}}0.647\\ {[}0.645, 0.649{]}\end{tabular}}} & \multicolumn{1}{l|}{{\color[HTML]{000000} \begin{tabular}[c]{@{}l@{}}0.703\\ {[}0.701, 0.704{]}\end{tabular}}} & \multicolumn{1}{l|}{{\color[HTML]{000000} \begin{tabular}[c]{@{}l@{}}0.642\\ {[}0.640, 0.644{]}\end{tabular}}} & {\color[HTML]{000000} \begin{tabular}[c]{@{}l@{}}0.687\\ {[}0.685, 0.688{]}\end{tabular}} \\ \hline
\multicolumn{1}{|l|}{{\color[HTML]{000000} \textit{\begin{tabular}[c]{@{}l@{}}External\\ (MIMIC)\end{tabular}}}} & \multicolumn{1}{l|}{{\color[HTML]{000000} 6M}} & \multicolumn{1}{l|}{\cellcolor[HTML]{67FD9A}{\color[HTML]{000000} \begin{tabular}[c]{@{}l@{}}0.669\\ {[}0.658, 0.677{]}\end{tabular}}} & \multicolumn{1}{l|}{\cellcolor[HTML]{67FD9A}{\color[HTML]{000000} \begin{tabular}[c]{@{}l@{}}0.677\\ {[}0.662, 0.690{]}\end{tabular}}} & \multicolumn{1}{l|}{{\color[HTML]{000000} \begin{tabular}[c]{@{}l@{}}0.623\\ {[}0.619, 0.642{]}\end{tabular}}} & \multicolumn{1}{l|}{{\color[HTML]{000000} \begin{tabular}[c]{@{}l@{}}0.638\\ {[}0.620, 0.649{]}\end{tabular}}} & \multicolumn{1}{l|}{\cellcolor[HTML]{67FD9A}{\color[HTML]{000000} \begin{tabular}[c]{@{}l@{}}0.640\\ {[}0.627, 0.650{]}\end{tabular}}} & \cellcolor[HTML]{67FD9A}{\color[HTML]{000000} \begin{tabular}[c]{@{}l@{}}0.710\\ {[}0.685, 0.713{]}\end{tabular}} \\ \hline
\multicolumn{1}{|l|}{{\color[HTML]{000000} \textit{\begin{tabular}[c]{@{}l@{}}External\\ (ED)\end{tabular}}}} & \multicolumn{1}{l|}{{\color[HTML]{000000} 1yr}} & \multicolumn{1}{l|}{{\color[HTML]{000000} \begin{tabular}[c]{@{}l@{}}0.737\\ {[}0.734, 0.741{]}\end{tabular}}} & \multicolumn{1}{l|}{{\color[HTML]{000000} \begin{tabular}[c]{@{}l@{}}0.810\\ {[}0.806, 0.813{]}\end{tabular}}} & \multicolumn{1}{l|}{\cellcolor[HTML]{67FD9A}{\color[HTML]{000000} \begin{tabular}[c]{@{}l@{}}0.772\\ {[}0.767, 0.773{]}\end{tabular}}} & \multicolumn{1}{l|}{\cellcolor[HTML]{67FD9A}{\color[HTML]{000000} \begin{tabular}[c]{@{}l@{}}0.837\\ {[}0.833, 0.839{]}\end{tabular}}} & \multicolumn{1}{l|}{{\color[HTML]{000000} \begin{tabular}[c]{@{}l@{}}0.673\\ {[}0.671, 0.680{]}\end{tabular}}} & {\color[HTML]{000000} \begin{tabular}[c]{@{}l@{}}0.753\\ {[}0.748, 0.761{]}\end{tabular}} \\ \hline
\end{tabular}%
}
\end{table}
For multimodal models, MOSCARD model with causality achieves the best performance on both internal and external datasets and outperformed MedCLIP (contrastive) and ALBEF (knowledge distillation) baselines (Table~\ref{table:multi_modality}). Each multimodal model produces three types of predictions: (1) `combined' is derived from the concatenation of CXR and ECG features after alignment, (2) `CXR' and (3) `ECG' solely on Chest X-ray and ECG features from the aligned space. The best performance on the internal dataset is observed in `combined' predictions from MOSCARD, particularly for MACE prediction at 1 year, with an AUC of 0.750. Interestingly, we found that with MOSCARD, the predictions based solely on ECG features consistently outperform those from the single modality ECG models, highlighting benefits of multimodal integration in improving ECG-based predictions even with single directional attention. 

\begin{figure}[tb!]
    \centering
    \includegraphics[width=0.9\linewidth]{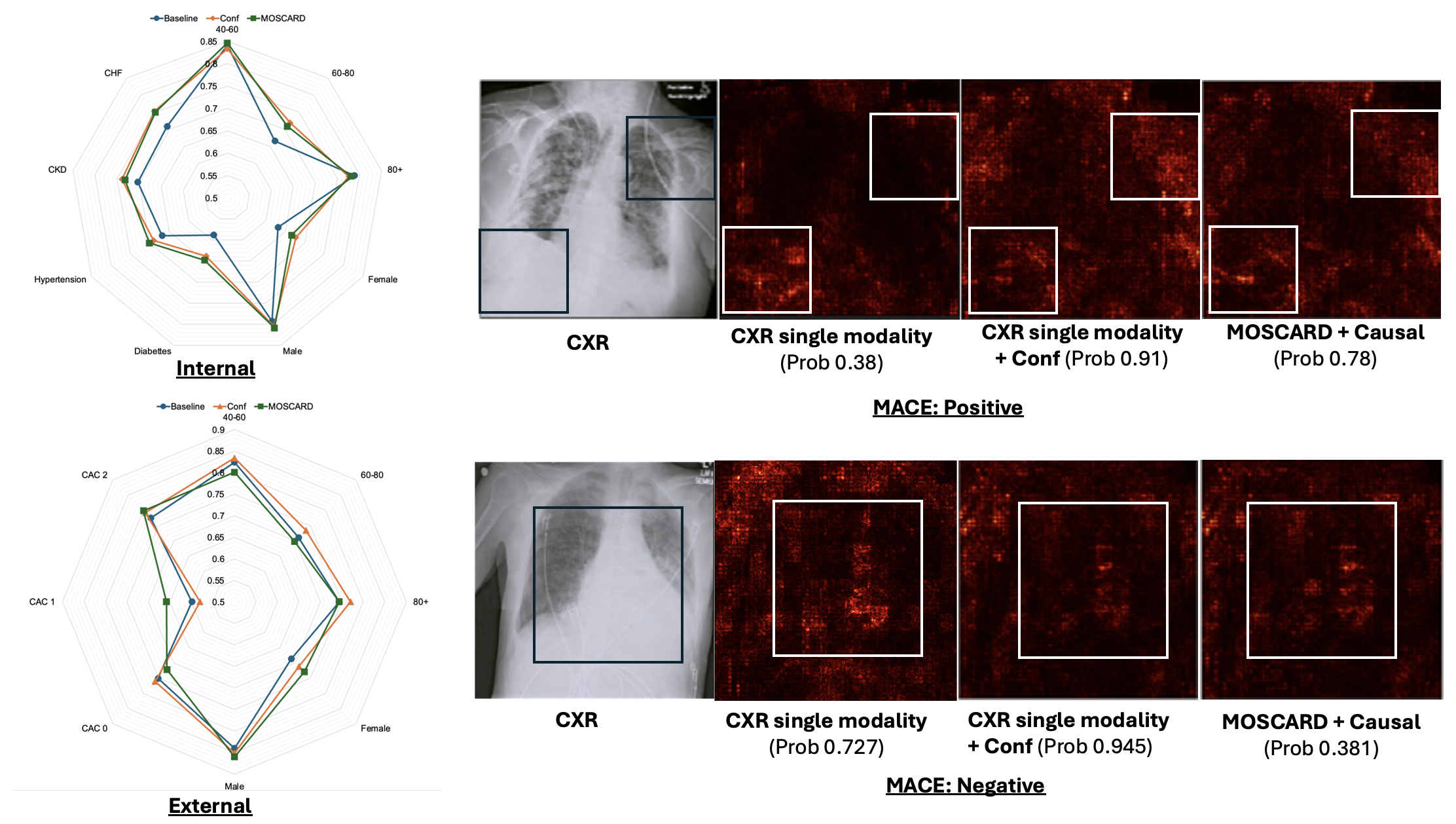}
    \caption{\footnotesize Spoke plot shows the comparative assessment of MOSCARD (green) against the multimodal co-alignment (blue) and with encoder after de-confounding (orange). Qualitative analysis with saliency maps - (a) Older male patient, true MACE event but single modality failed to predict (prob 0.38), MOSCARD correctly predicted (prob 0.78); (b) Young male patient, no MACE event, single and confounder model predicted false positive but proposed model correctly predicted (prob 0.38).}
    \label{S-map}
\end{figure}

MOSCARD with causality (CXR branch) achieves the best performance on external ED dataset(AUC = 0.837), while MODSCARD (ECG branch) performs best on MIMIC (AUC = 0.71) for MACE prediction at 6 months. These findings highlight how dataset-specific characteristics impact modality effectiveness; however, similar to the internal datasets, multimodal learning enhances performance on both the external ED and MIMIC datasets. Our results indicate that for internal datasets, MOSCARD with the confounder model achieves the best performance against the single modality. Meanwhile, for external datasets, MOSCARD outperforms other models which suggest that de-confounding and causal learning are particularly beneficial for external datasets, likely due to differences in disease prevalence and demographic distributions. Fig.~\ref{S-map} shows that MOSCARD achieved optimal equalized performance across all subgroups including comorbidity and demographics for both internal and external ED data. Deconfounded encoders improved the subgroup performance from the baseline and achieved similar performance on the hold-out but the performance drop on the external ED test. Interestingly, even though CT calcium score (CT imaging marker for coronary atherosclerosis) was not used as a factor for causal intervention, MOSCARD also achieved optimal performance across all the calcium subgroups. 


In the context of two-step learning, we applied saliency map interpretation with pixel masking on several interesting cases to explore how MOSCARD reasoning differs from single modality encoding (Fig~\ref{S-map}). In both cases, the single modality model mainly relied on external wires and support devices for making predictions. Despite after de-confounding, the focus shifted, but the model's predictions remained either incorrect or overly confident. In contrast, with ECG guidance, MOSCARD successfully made the correct predictions in both cases, focusing on the heart and lung regions without being distracted by external wires and support devices.

\section{Discussion}
We introduce the MOSCARD, a novel multimodal causal reasoning model designed to support opportunistic screening of MACE by leveraging existing CXR and 12-lead ECG data, while addressing the challenges posed by causality and confounding factors. Experimental results highlight the effectiveness of MOSCARD in providing more accurate and fair predictions for shift in patient populations (high to low risk), compared to traditional single modality and existing multimodal models. De-confounding showed its strength in improving generalizability on external datasets, with the CXR Conf achieving an AUC of 0.778 on the ED dataset and ECG Conf reaching 0.66 AUC on the MIMIC dataset. The MOSCARD with causal reasoning achieved the best performance on external datasets, with the causal version reaching an AUC of 0.837 for MACE at 1 year on the ED dataset. For MIMIC, MOSCARD's ECG branch performed better (AUC = 0.71). MOSCARD also achieved optimal performance across all subgroups, including those based on comorbidities and demographics, for both internal and external ED datasets.

Rather than relying on patients presenting specifically for cardiovascular evaluations, we propose a novel application for the secondary utilization of multimodal clinical data to enable the early identification of individuals at risk for MACE. This early detection strategy not only facilitates timely interventions and lifestyle modifications but also holds promise in resource-limited settings. On the technical front, our primary contribution is the development of a generalizable multimodal prognostic framework designed to perform robustly across diverse, previously unseen populations with varying prevalence of cardiovascular events. Unlike much of the existing literature, which often overlooks the role of confounders and underlying causal structures in multimodal data, our approach explicitly integrates causal reasoning by incorporating comorbidities and mitigating demographic biases using confusion loss. This debiasing strategy enhances the model's generalizability and effectiveness across heterogeneous populations. The ED dataset includes healthier patients with fewer comorbidities and a lower MACE rate, whereas the MIMIC-IV represents a more complex ICU population with additional comorbidities. After co-alignment, the CXR branch performs best on the ED dataset by identifying subtle signs of early-stage cardiovascular abnormalities, such as early atherosclerosis or pulmonary congestion, while the ECG branch outperforms on the MIMIC-IV patients who are more likely to present with advanced stages of disease based on assessment of electrical activity; however, the co-alignment of CXR and ECG within MOSCARD can learn from both modalities and adjusting for the differences in the populations which help the model to significantly surpassed the performance of single modality. Additionally, our training process is divided into two steps. If one modality is unavailable, the model can still be evaluated using only the first step, which includes confounder debiasing. Due to the limited availability of open-source global datasets with MACE labels, we were unable to assess the model’s generalizability beyond the U.S. healthcare setting, which represents a limitation of our study. 

\begin{credits}
\subsubsection{\ackname} The work is supported by NHLBI funded proposal titled \emph{Opportunistic Screening for ASCVD using a Multimodal Deep Learning Risk Prediction Model} grant no. R01 HL167811-02 (PI: Banerjee).

\subsubsection{\discintname}
The authors have no competing interests to declare that are
relevant to the content of this article. 

\end{credits}

%
%
%
\bibliographystyle{splncs04}
\bibliography{output}

\end{document}